\documentclass[runningheads]{llncs}

\usepackage{eccv}

\usepackage{eccvabbrv}

\usepackage{graphicx}
\usepackage{booktabs}

\usepackage[accsupp]{axessibility}  

\usepackage[pagebackref,breaklinks,colorlinks,citecolor=eccvblue]{hyperref}

\usepackage{orcidlink}

\begin{document}

\title{Scalable Diffusion Models with State Space Backbone} 

\titlerunning{DiS}

\author{%
  {Zhengcong Fei}, {Mingyuan Fan}, Changqian Yu\\{Junshi Huang*} \\
  \institute{Kunlun Inc.} 
  \email{\{feizhengcong\}@gmail.com} 
}
\authorrunning{Fei et al.}

\maketitle

\begin{abstract}
  This paper presents a new exploration into a category of diffusion models built upon state space architecture. We endeavor to train diffusion models for image data, wherein the traditional U-Net backbone is supplanted by a state space backbone, functioning on raw patches or latent space. Given its notable efficacy in accommodating long-range dependencies, Diffusion State Space Models (DiS) are distinguished by treating all inputs including time, condition, and noisy image patches as tokens. Our assessment of DiS encompasses both unconditional and class-conditional image generation scenarios, revealing that DiS exhibits comparable, if not superior, performance to CNN-based or Transformer-based U-Net architectures of commensurate size. Furthermore, we analyze the scalability of DiS, gauged by the forward pass complexity quantified in Gflops. DiS models with higher Gflops, achieved through augmentation of depth/width or augmentation of input tokens, consistently demonstrate lower FID. In addition to demonstrating commendable scalability characteristics, DiS-H/2 models in latent space achieve performance levels akin to prior diffusion models on class-conditional ImageNet benchmarks at the resolution of 256$\times$256 and 512$\times$512, while significantly reducing the computational burden. The code and models are available at: 
\url{https://github.com/feizc/DiS}. 
\end{abstract}

\section{Introduction} 

Diffusion models \cite{ho2020denoising,sohl2015deep,song2020score,cao2024survey} 
have emerged as potent deep generative models in recent years
\cite{dhariwal2021diffusion,ho2022cascaded,rombach2022high}, for their capacity in high-quality image generation. 
Their rapid development has led to widespread application across various domains, including text-to-image generation \cite{ramesh2022hierarchical,rombach2022high,saharia2022photorealistic,gu2022vector}, image-to-image generation \cite{choi2021ilvr,zhao2022egsde,zhang2023adding}, video generation \cite{ho2022imagen,mei2023vidm,mei2023vidm},
speech synthesis \cite{chen2020wavegrad,kong2020diffwave}, and 3D synthesis \cite{poole2022dreamfusion,zhou2023sparsefusion}.
Along with the development of sampling algorithms \cite{dockhorn2021score,kingma2021variational,lu2022maximum,lu2022dpm,song2021maximum,vahdat2021score}, the revolution of backbones stands as a pivotal aspect in advancement of diffusion models. 
A representative example is U-Net based on a convolutional neural network (CNN) \cite{ho2020denoising,song2019generative}, which has been prominently featured in previous research. 
The CNN-based UNet is characterized by a group of down-sampling blocks, a group of up-sampling blocks, and long skip connections between the two groups \cite{dhariwal2021diffusion,ramesh2022hierarchical,rombach2022high,saharia2022photorealistic}. Similarly, Transformer-based architecture replaces sampling block with self-attention while keeping the remain untouched \cite{peebles2023scalable,yang2022your,bao2023all}, resulting in streamlined yet effective performance.

On the other hand, state space models (SSMs) with efficient hardware-aware designs, have shown great potential in the realm of long sequence modeling \cite{kalman1960new,gu2021efficiently,gu2021combining,gupta2022diagonal,orvieto2023resurrecting,smith2022simplified}. As self-attention mechanism in Transformer scales quadratically with the input size, making them resource-intensive when dealing with long-range visual dependencies, \emph{i.e.}, high resolution images. 
Recent efforts, exampified by the work on Mamba \cite{gu2023mamba,fu2022hungry}, have sought to address by integrating time-varying parameters into the SSM and proposing a hardware-aware algorithm to enable efficient training and inference. 
The commendable scaling performance of Mamba underscores its promise as a viable avenue for constructing efficient and versatile backbones within the domain of SSMs. 
Motivated by the successes observed in language modeling with Mamba, a pertinent inquiry arises: whether we can build SSM-based U-Net in diffusion models?

\begin{figure}[t]
  \centering
   \includegraphics[width=1.\linewidth]{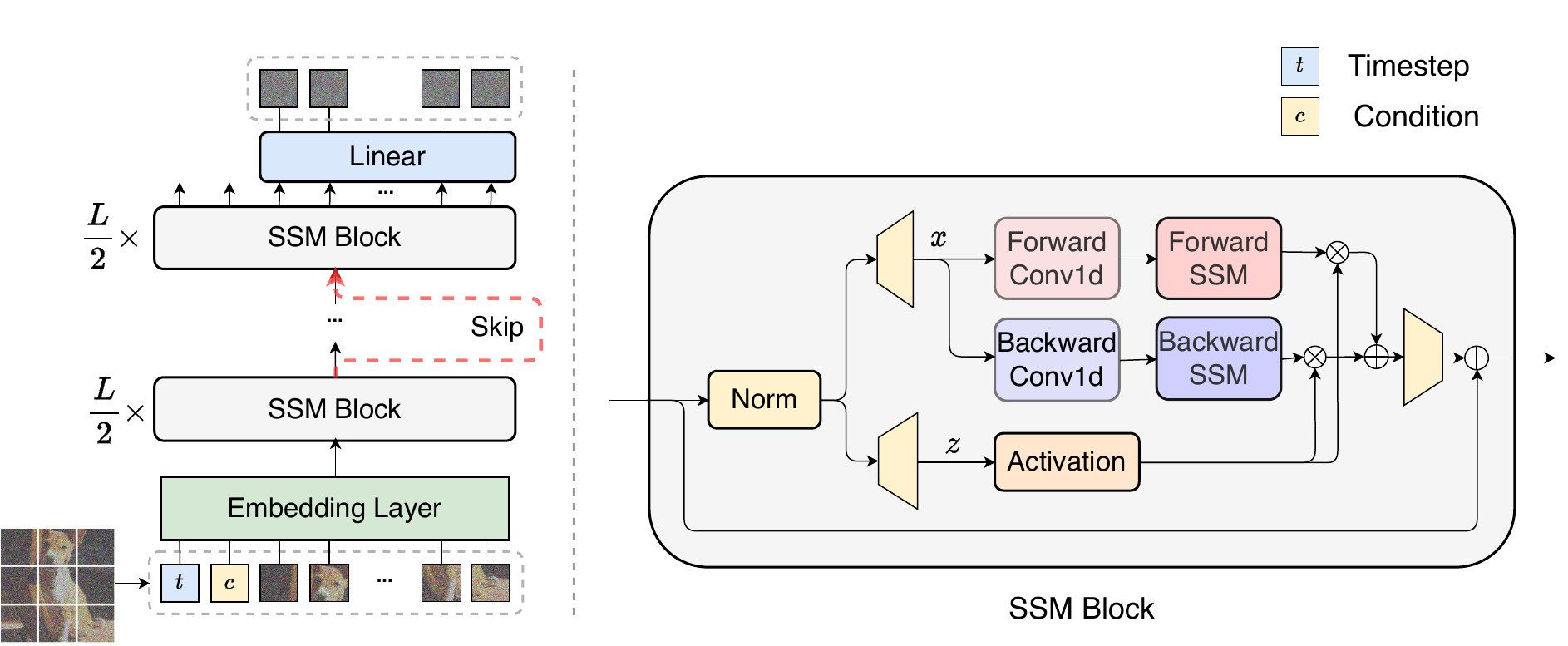}
   \caption{\textbf{The proposed state space-based diffusion models.} It treats all inputs including the time, condition and noisy image patches as tokens and employs skip connections between shallow and deep layers. Different from original Mamba for text sequence modeling, our SSM block process the hidden states sequence with both forward and backward directions. }
   \label{fig:framework} 
\end{figure}

In this paper, we aim to design a simple and general state space-based architecture for diffusion models, referred to as DiS. Following the design principles of \cite{peebles2023scalable}, DiS treats all inputs including time, condition and noisy image patches as discrete tokens. 
Crucially, we conduct a systematical ablation study encompassing the incorporation of conditioning factors and the optimization of model architecture across all components. Noteworthy is DiS's adherence to the established best practices of state space models, renowned for their superior scalability in image generation tasks when compared to CNN or Transformers, all while maintaining lower computational overhead.

Experimentally, we assess the performance DiS across both unconditional and class-conditional image generation tasks.  In all settings, DiS demonstrate comparative, if not superior, efficay when juxtaposed with CNN-based or Transformer-based U-Nets of a similar size. 
Additionally, we provide evidence showcasing that DiSs are scalable architectures for diffusion models, where a discernible correlation is observed between network complexity and sample quality. 
Moreover, experiments yield impressive results, with DiS achieving a comparable FID scores in class-conditional image generation conducted on ImageNet at a resolution of 256$\times$256 and 512$\times$512. 
This study endeavors to elucidate the criticality of architectural selections within the domain of diffusion models, concurrently providing empirical benchmarks that can inform forthcoming research in generative modeling. 
Our aspiration is that the insights gleaned from DiS will serve as foundational knowledge for future investigations concerning backbone architectures in diffusion models, ultimately enriching the landscape of generative modeling, particularly in the context of large-scale cross-modality datasets.

\section{Methodology}

\subsection{Preliminaries} 


\subsubsection{Diffusion models.}

Prior to presenting our architecture, we provide a concise overview of fundamental concepts relevant to our discussion. 
Diffusion models \cite{ho2020denoising,sohl2015deep} gradually inject noise to data, and subsequently reverse this process to generate data from noise. 
The noise injection process, also referred to as the forward process, can be formalized as Markov chain as:
\begin{equation}
    \begin{split}
        q(\textbf{x}_{1:T}|\textbf{x}_0) = \prod_{t=1}^{T}q(\textbf{x}_t|\textbf{x}_{t-1}),\\
        q(\textbf{x}_t|\textbf{x}_{t-1}) = \mathcal{N}(\textbf{x}_t| \sqrt{\alpha_t}\textbf{x}_{t-1}, \beta_t \textbf{I}),
    \end{split}
\end{equation}
Here $\textbf{x}_0$ is the data, $\alpha_t$ and $\beta_t$ represent the noise schedule, ensuring $\alpha_t + \beta_t = 1$. 
To reverse this process, a Gaussian model $p(\textbf{x}_{t-1}|x_t) = \mathcal{N}(\textbf{x}_{t-1}| \mu_t(\textbf{x}_t), \sigma_t^2 \textbf{I} )$ is employed to approximate the ground truth reverse transition $q(\textbf{x}_{t-1}|\textbf{x}_t)$, where the learning is tantamount to a noise prediction task. 
Formally, a noise prediction network $\epsilon_\theta (\textbf{x}_t, t)$ is incorporated by minimizing a noise prediction objective, \emph{i.e.}, $\min_\theta \mathbb{E}_{t, \textbf{x}_0, \epsilon} || \epsilon - \epsilon_\theta (\textbf{x}_t, t)||_2^2 $, where $t$ is uniformly distributed between 1 and $T$. 
To learn conditional diffusion models, \emph{e.g.}, class-conditional \cite{dhariwal2021diffusion} or text-to-image \cite{ramesh2022hierarchical,betker2023improving} models, additional condition information is integrated into the noise prediction objective as: 
\begin{equation}
    \min_\theta \mathbb{E}_{t, \textbf{x}_0, \textbf{c}, \epsilon} || \eta - \eta_\theta(\textbf{x}_t, t, \textbf{c}) ||^2_2,
\end{equation}
where $\textbf{c}$ is the condition index or its continuous embedding.

\subsubsection{State space backbone.} 
State space models are conventionally defined as linear time-invariant systems that builds a map stimulation $x(t) \in \mathbb{R}^N$ to response $y(t) \in \mathbb{R}^N$ by a latent state $h(t)\in \mathbb{R}^N$. The process can be formulated as: 
\begin{equation}
\begin{aligned}
    h'(t) &= \textbf{A}h(t) + \textbf{B}x(t), \\
    y(t) &= \textbf{C}h(t),
\end{aligned}
\label{eq:ssm-con}
\end{equation}
where $\textbf{A}\in \mathbb{R}^{N\times N}$ and $\textbf{B}, \textbf{C}\in \mathbb{R}^N$ denote the state matrix, input matrix, and output matrix, respectively. 
In the quest to derive the output sequence $y(t)$ at time $t$, the analytical solution for obtaining $h(t)$ proves to be a formidable challenge. Conversely, real-world data typically manifests in discrete rather than continuous forms. As an alternative, we can discretize the system in Equation \ref{eq:ssm-con} as follows:
\begin{equation}\label{SSM}
\begin{split}
    h_{t} & = \overline{\textbf{A}}h_{t-1} + \overline{\textbf{B}}x_{t}, \\
    y_{t} & = \textbf{C} h_{t},
\end{split}
\end{equation}
where $\overline{\textbf{A}}:=\exp(\Delta \cdot \textbf{A})$ and $\overline{\textbf{B}}:=(\Delta \cdot \textbf{A})^{-1}(\exp(\Delta\cdot \textbf{A})-I)\cdot \Delta \textbf{B}$ are the discretized state parameters and $\Delta$ is the discretization step size. 

While SSMs boast rich theoretical properties, they are often afflicted by high computational costs and numerical instability. These shortcomings have prompted the development of structured state space sequence models (S4), which seek to mitigate these challenges by imposing a structured format on the state matrix $\textbf{A}$ through the utilization of HIPPO matrices. This structural enhancement has led to notable advancements in both performance and efficiency. Notably, S4 has exhibited a substantial performance advantage over Transformers, which necessitates effective modeling of long-range dependencies \cite{gu2021efficiently}. More recently, Mamba \cite{gu2023mamba} further advance its potential through an input-dependent selection mechanism and a faster hardware-aware algorithm.

\subsection{Model Structure Design} 

We introduce Diffusion State Space Models, denoted as DiS, a simple and general architecture for diffusion models in image generation.
Specifically, DiS parameterizes the noise prediction network $\epsilon_\theta(\textbf{x}_t, t, \textbf{c})$, which takes the timestep $t$, condition $\textbf{c}$ and noised image $\textbf{x}_t$ as inputs and predicts the noise injected into $\textbf{x}_t$. 
Our objective is to closely adhere to the advanced state space architecture to preserve its scalability characteristics, hence, DiS is based on the bidirectional Mamba \cite{gu2023mamba,zhu2024vision} architecture which operates on sequences of tokens. Figure \ref{fig:framework} provides an overview of the complete DiS architecture. In this section, we expound upon the forward pass of DiS, as well as the components of the design space of the DiS class.

\subsubsection{Image patchnify.}
The first layer of DiS undertakes transformation of input image $\textbf{I} \in \mathbb{R}^{H \times W \times C}$ into flatten 2-D patches $\textbf{X} \in \mathbb{R}^{J \times (p^2 \cdot C)}$. 
Subsequently, converts it into a sequence of $J$ tokens, each of dimension $D$, by linearly embedding each patch in the input. Consistent with \cite{dosovitskiy2020image}, we apply learnable positional embeddings to all input tokens.
The number of tokens $J$ created by patchify is determined by the hyperparameter patch size $p$ to $\frac{H \times W}{p^2}$. The patchnify supports both raw pixel and latent space.
We set $p=2, 4, 8$ to the design space.

\subsubsection{SSM block.}
After embedding layer, the input tokens are processed by a sequence of SSM blocks. In addition to noised image inputs, diffusion models sometimes process additional conditional information such as noise timesteps $t$, condition $\textbf{c}$ such as class labels or natural language. 
Given that original Mamba block is designed for 1-D sequence, we resort to \cite{zhu2024vision}, which incorporates bidirectional sequence modeling tailed for vision tasks. 
The designs introduce subtle yet pivotal modifications to the standard SSM block design. As shown in Figure \ref{fig:framework} right part, forward of SSM blocks combine both forward and backward directions.

\subsubsection{Skip connection.}
Given a series of $L$ SSM blocks, we categorize the stack SSM block into first half $\lfloor \frac{L}{2} \rfloor$ shallow group, one middle layer, and second half $\lfloor \frac{L}{2} \rfloor$ deep group. Let $\textbf{h}_{shallow}, \textbf{h}_{deep} \in \mathbb{R}^{J \times D}$ be the hidden states from the main branch and long skip branch respectively. We consider directly concatenating them and perform a linear projection, \emph{i.e.}, \texttt{Linear(Concat(}$\textbf{h}_{shallow}, \textbf{h}_{deep}$\texttt{))}, before feeding them to the next SSM block.

\subsubsection{Linear decoder.}
After the final SSM block, we need to decode our sequence of hidden states into an output noise prediction and diagonal covariance prediction.
Both of these outputs retain a shape identical to the original spatial input. 
We use a standard linear decoder, \emph{i.e.}, we apply the final layer norm and linearly decode each token into a $p^2 \cdot C$ tensor. Finally, we rearrange the decoded tokens into their original spatial layout to get the predicted noise and covariance.

\subsubsection{Condition incorporation.}
To effectively integrate additional conditions, we adopt a straightforward strategy of appending the vector embeddings of timestep $t$ and condition $\textbf{c}$ as two supplementary tokens in the input sequence. 
These tokens are treated equivalently to the image tokens, akin to the approach with cls tokens in Vision Transformers \cite{dosovitskiy2020image}. This approach enables the utilization of SSM blocks without necessitating any modifications.  
Following the final block, the conditioning tokens are removed from the sequence.
We also explore to adaptive normalization layer, where replace standard norm layer with adaptive norm layer. That is, rather than directly learn dimension-wise scale and shift parameters, we regress them from the sum of the embedding vectors of $t$ and $\textbf{c}$, which will be discussed in the Experiments Section.

\begin{table}[t]
\caption{\textbf{Scaling law model size.} The model sizes and detailed hyperparameters settings for scaling experiments. 
}
\centering
\setlength{\tabcolsep}{1.5mm}{
\begin{tabular}{lccccc}
\toprule
& \#Params & \#Blocks $L$  & Hidden dim. $D$ & \#Expand $E$ &Gflops  \\ \midrule
Small  & 28.4M &25 & 384 &2& 0.43 \\
Base &119.1M&25 &768&2& 1.86 \\
Medium &229.4M &49 & 768 &2 &3.70  \\
Large &404.0M&49&1024&2&6.57 \\
Huge &900.6M&49 &1536&2& 14.79 
\\ \bottomrule 
\end{tabular}}
\label{tab:scale}
\end{table}

\subsection{Computation Analysis}

In summary, the hyper-parameters of our architecture encompass the following: 
the number of blocks $L$, hidden state dimension $D$, expanded state dimension $E$, and SSM dimension $N$. 
Various configurations of DiS are delineated in Table \ref{tab:scale}. They cover a wide range of model sizes and flop allocations, from 28M to 900M, thus affording comprehensive insights into the scalability performance.
Aligned with \cite{peebles2023scalable}, Gflop metric is evaluated in 32$\times$32 unconditional image generation with patch size $p=4$ with \texttt{thop} python package.
Following \cite{gu2023mamba} we also set the SSM dimension of all model variants to 16.

Both SSM block in DiS and self-attention in Transformer play a key role in modeling long context adaptively. 
We further provide a theoretical analysis pertaining to computation efficiency.
Given a sequence $\textbf{X} \in \mathbb{R}^{1 \times J \times D}$ and the default setting $E=2$, the computation complexity of a self-attention and SSM operation are delineated as: 
\begin{align}
    \mathcal{O}(\text{SA}) = 4JD^2 + 2J^2D,\\
    \mathcal{O}(\text{SSM}) = 3J(2D)N + J(2D)N^2,
\end{align}
where we can see that self-attention is quadratic to sequence length $J$, and SSM is linear with respect to sequence length $J$.  It is noteworthy that $N$ is a fixed parameter, typically set to 16 by default. This computational efficiency renders DiS amenable to scalability in scenarios necessitating generation with large sequence lengths, such as gigapixel applications.

\section{Experiments}

We delve into the design space and scrutinize the scaling properties of our DiS model class. Our model is named according to their configs and patch size $p$; for instance, DiS-L/2 refers to the Large version config and $p=2$. 

\subsection{Experimental Settings}

\subsubsection{Datasets.}
For unconditional image learning, we consider CIFAR10 \cite{krizhevsky2009learning}, which comprises 50K training images, and CelebA 64x64 \cite{liu2015deep}, which contains 162,770 training images of human faces. For class-conditional image learning, we consider ImageNet \cite{deng2009imagenet} at 256$\times$256 and 512$\times$512 resolutions, which contains 1,281,167 training images across 1,000 different classes. 
The only data augmentation is horizontal flips. 
We train 500K iterations on CIFAR10 and CelebA 64$\times$64 with a batch size of 128. We train 500K and 1M iterations on ImageNet 256$\times$256 and 512$\times$512, with a batch size of 1024. 

\subsubsection{Implementation details. }
We use the AdamW optimizer \cite{kingma2014adam} without weight decay across all datasets. 
We maintain a learning rate of 1e-4 with a cosine schedule. 
In our early experiments, we tried learning rates ranging from 1e-4 to 5e-4, and found that loss may become NAN if the learning rate set is larger than 3e-4. 
Consistent with practice in literature \cite{peebles2023scalable}, we utilize an exponential moving average of DiS weights over training with a decay of 0.9999. All results were reported using the EMA model. Our models are trained on Nvidia A100 GPU.  
When trained on ImageNet dataset at resolution of 256$\times$256 and 512$\times$512, we adopt classifier-free guidance \cite{ho2022classifier} following \cite{rombach2022high} and use a off-the-shelf pre-trained variational autoencoder (VAE) model \cite{kingma2013auto} from Stable Diffusion \cite{rombach2022high} provided in huggingface\footnote{https://huggingface.co/stabilityai/sd-vae-ft-ema}.
The VAE encoder has a downsampling factor of 8.
We retrain diffusion hyperparameters from \cite{peebles2023scalable}, using a $t_{max}=1000$ linear variance schedule ranging from $1\times10^{-4}$ to $2\times10^{-2}$ and parameterization of the covariance.

\begin{figure}[t]
  \centering
   \includegraphics[width=1.\linewidth]{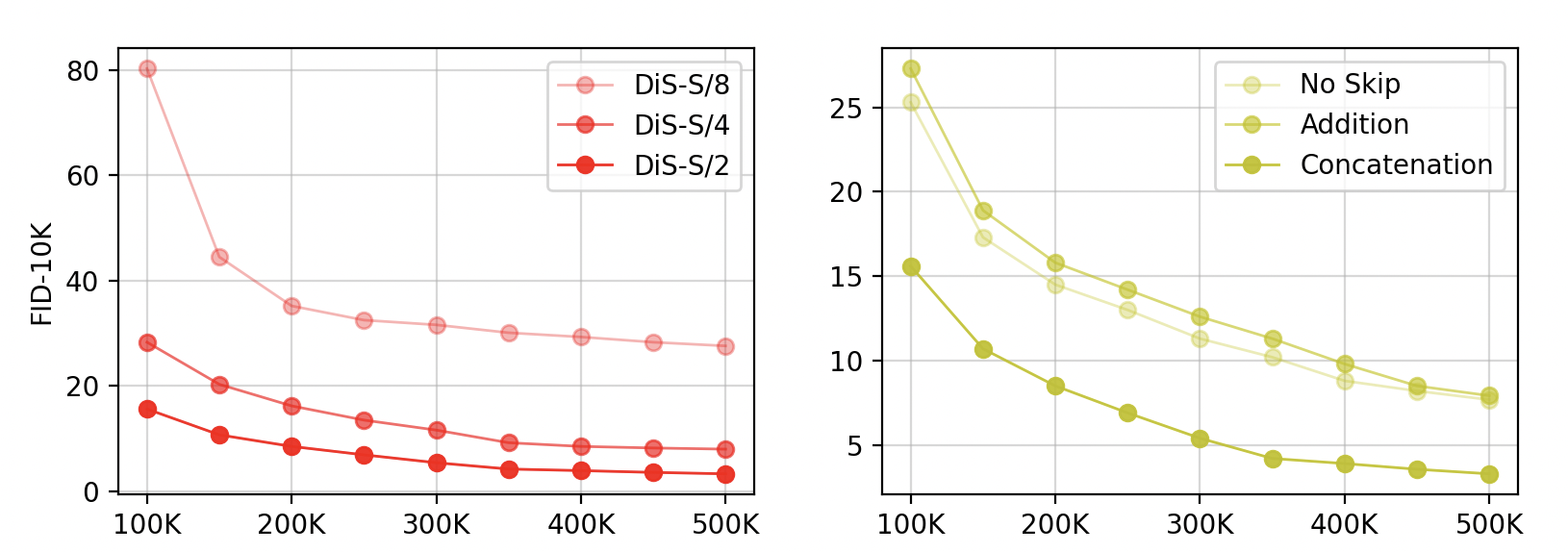}
   \caption{\textbf{Ablation experiments} with DiS-S/2 on CIFAR10 dataset. We report FID metrics on 10K generated samples. (a) \textbf{Patch size}. A smaller patch size can improve the generation performance. (b) \textbf{Long skip}. Combining the long skip branch can accelerate the training as well as optimize generated results. 
   }
   \label{fig:ab1} 
\end{figure}

\subsubsection{Evaluation metrics.}
We measure image generation performance with Frechet Inception Distance (FID) \cite{heusel2017gans}, a widely adopted metric for assessing the quality of generated images.  We follow convention when comparing against prior works and report FID-50K using 250 DDPM sampling steps \cite{parmar2022aliased} following the process of \cite{dhariwal2021diffusion}. Notably, the FID values presented in this section are calculated without classifier-free guidance, unless specified otherwise. We additionally report Inception Score \cite{salimans2016improved}, sFID \cite{nash2021generating} and Precision/Recall \cite{kynkaanniemi2019improved} as secondary metrics.

\subsection{Model Analysis}
In this section, we perform a systematical empirical investigation into the fundamental components of DiS models. Specifically, we ablate on the CIFAR10 dataset, evaluate the FID score every 50K training iterations on 10K generated samples, instead of 50K samples for efficiency identical to \cite{bao2023all}, and determine the optimal default implementation details.

\subsubsection{Effect of patch size.}

We train patch size range over (8, 4, 2) with DiS-S configuration on the CIFAR10 dataset. 
As depicted in Figure \ref{fig:ab1}(a) shows FID metric fluctuates in response to patch size decrease when maintaining model size consistent. 
We observe that discernible FID enhancements are observed throughout the training process by augmenting the number of tokens processed by DiS, holding parameters approximately fixed. 
Consequently, optimal performance necessitates a smaller patch size, such as 2. We postulate that this requirement stems from the inherently low-level nature of the noise prediction task in diffusion models, which favors smaller patches, in contrast to higher-level tasks like classification. Moreover, given the computational costs associated with using small patch sizes for high-resolution images, an alternative approach involves transforming these images into low-dimensional latent representations, subsequently also modeled using the DiS series.

\begin{figure}[t]
  \centering
   \includegraphics[width=1.\linewidth]{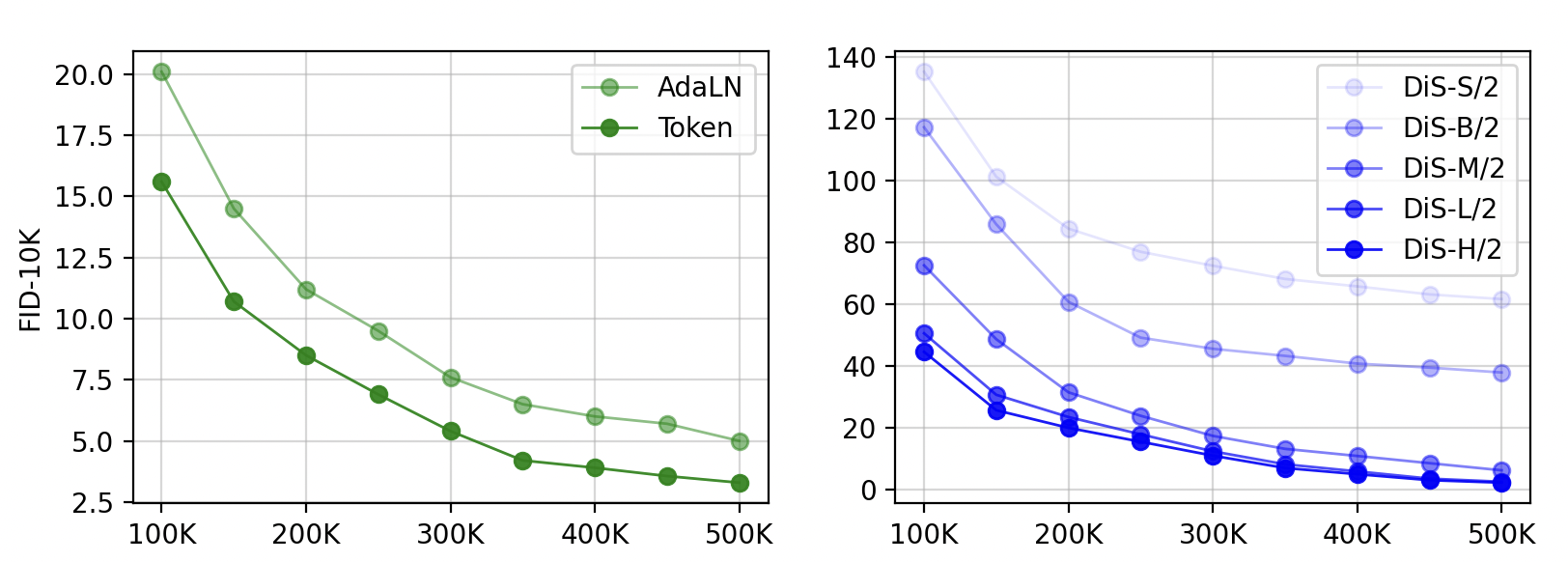}
   \caption{\textbf{Model analysis for different designs.} (a) \textbf{Variants of condition incorporation}. Feed time as token into the networks is effective. (b) \textbf{Model parameters scaling}. As we expected, holding patch size constant, increasing the model size can consistently improve the generation performance. 
   }
   \label{fig:ab2} 
\end{figure}

\subsubsection{Effect of long skip.}

For evaluating the efficiency of skipping operation, we examine three variants including: ($\textbf{i}$) Concatenation, \emph{i.e.}, \texttt{Linear(Concat(}$\textbf{h}_{shallow}$, $\textbf{h}_{deep}$\texttt{))}; ($\textbf{ii}$) Addition, \emph{i.e.}, $\textbf{h}_{shallow} + \textbf{h}_{deep}$; ($\textbf{iii}$) No skip connection. 
As show in Figure \ref{fig:ab1}(b), directly adding the hidden states from the shallow and deep layers does not yield any discernible benefits. Given that the SSM block already incorporates residual skip connections internally, the hidden states of deep layer inherently encapsulates some information from the shallow layer in a linear fashion. Consequently, the addition operation in the long skip merely amplifies the contribution of the shallow layer's hidden state, without fundamentally altering the network's behavior. In contrast, employing concatenation involves a learnable linear projection on the shallow hidden states and effectively enhances performance compared to the absence of a long skip connection.

\begin{figure}[!t]
\begin{minipage}
{\textwidth}
    \begin{minipage}[t]{0.45\textwidth}
            \centering
            \small \makeatletter\def\@captype{table}\makeatother\caption{\textbf{Benchmarking unconditional image generation on CIFAR10}. DiS-S/2 model obtains comparable results with fewer parameters. }
            \vspace{0.3cm}
            \setlength{\tabcolsep}{1.mm}{
            \begin{tabular}{lcc}
\toprule 
\multicolumn{3}{l}{\textbf{Unconditional CIFAR10}} \\
\midrule
Model    &\#Params &FID$\downarrow$   \\ \midrule
\multicolumn{3}{l}{ \textcolor{gray}{\emph{Diff. based on U-Net}}}\\
DDPM \cite{ho2020denoising} &36M& 3.17 \\
EDM \cite{karras2022elucidating} &56M& 1.97 \\ 
\multicolumn{2}{l}{\textcolor{gray}{\emph{Diff. based on Transformer}}}\\
GenViT \cite{yang2022your} &11M& 20.20  \\
U-ViT-S/2 \cite{bao2023all} & 44M& 3.11 \\ 
\multicolumn{2}{l}{\textcolor{gray}{\emph{Diff. based on SSM}}}\\
DiS-S/2 &28M &3.25 \\
\bottomrule
\end{tabular}}
        \end{minipage}
        \hspace{.15in}
        \begin{minipage}[t]{0.45\textwidth}
    \centering    \small \makeatletter\def\@captype{table}\makeatother\caption{\textbf{Benchmarking unconditional image generation on CelebA 64$\times$64}. DiS-S/2 maintains a superior generation performance in small model settings.}
    \vspace{0.3cm}
        \setlength{\tabcolsep}{1.mm}{
            \begin{tabular}{lcc}
\toprule
\multicolumn{3}{l}{\textbf{Unconditional CelebA 64$\times$64}} \\ \midrule
Model &\#Params  &FID$\downarrow$ \\ \midrule
\multicolumn{2}{l}{ \textcolor{gray}{\emph{Diff. based on U-Net}}}\\
DDIM \cite{song2020denoising} &79M & 3.26    \\
Soft Trunc. \cite{kim2021soft} &62M & 1.90 \\
\multicolumn{2}{l}{\textcolor{gray}{\emph{Diff. based on Transformer}}}\\
U-ViT-S/4 \cite{bao2023all}& 44M & 2.87\\
\multicolumn{2}{l}{\textcolor{gray}{\emph{Diff. based on SSM}}}\\
DiS-S/2& 28M &2.05 \\
\bottomrule 
\end{tabular}}
\end{minipage}
\end{minipage}
\end{figure}

\subsubsection{Effect of condition combination.}
We explore two ways to integrate the conditional timestep $t$ into the network: ($\textbf{i}$) treat timestep $t$ as a token and directly concatenate it with image patches in prefix. ($\textbf{ii}$) incorporate the time embedding after the layer normalization in the SSM block, which is similar to the adaptive group normalization \cite{dhariwal2021diffusion} used in U-Net. 
The second way is referred to as adaptive layer normalization (AdaLN). Formally, AdaLN($h,s$)=$y_s \text{LayerNorm}(h) + y_b$, where $h$ is a hidden state inside an SSM block, and $y_s$ and $y_b$ are obtained from a linear projection of the time embedding. 
As illustrated in Figure \ref{fig:ab2} (a), while simple and straightforward, the first way that treats time as a token performs better than AdaLN.

\subsubsection{Scaling model size.}
We investigate scaling properties of DiS by studying the effect of depth, \emph{i.e.}, number of SSM layers, width, e.g. the hidden size. Specifically, we train 5 DiT models on the ImageNet dataset with a resolution of 256$\times$256, spanning model configurations from small to huge as detailed in Table \ref{tab:scale}, denoted as (S, B, M, L, H) for simple. 
As shown in Figure \ref{fig:ab2} (b), the performance improves as the depth increase from 25 to 49. Similarly, increasing the width from 384 to 768  yields performance gains. 
Notably, the model's performance has not yet reached convergence. 
Overall, across all five configurations, significant enhancements in FID are observed across all training stages by augmenting the depth and width of the SSM structure.

\subsection{Main Results}

\begin{table}[t]
\centering
\caption{\textbf{Benchmarking class-conditional image generation on ImageNet 256$\times$256.} DiS-H/2 achieves state-of-the-art FID metrics towards best competitors.} %
    \setlength{\tabcolsep}{2.5mm}{
    \begin{tabular}{lccccc}
    \toprule
    \multicolumn{6}{l}{\bf{Class-Conditional ImageNet} 256$\times$256} \\
     \midrule
    Model & FID$\downarrow$   & sFID$\downarrow$  & IS$\uparrow$     & Precision$\uparrow$ & Recall$\uparrow$ \\
      \midrule
    \multicolumn{6}{l}{ \textcolor{gray}{\emph{GAN}}} \\
    BigGAN-deep~\cite{brock2018large} & 6.95 & 7.36 & 171.4 & 0.87 & 0.28 \\
    StyleGAN-XL~\cite{sauer2022stylegan} & 2.30 & 4.02 & 265.12 & 0.78 & 0.53 \\
    \multicolumn{6}{l}{ \textcolor{gray}{\emph{Diff. based on U-Net}}}\\
    ADM~\cite{dhariwal2021diffusion} & 10.94 & 6.02 & 100.98 & 0.69 & 0.63 \\
    ADM-U & 7.49 & 5.13 & 127.49 & 0.72 & 0.63 \\
    ADM-G & 4.59 & 5.25 & 186.70 & 0.82 & 0.52 \\
    ADM-G, ADM-U & 3.94 & 6.14      & 215.84 & 0.83 & 0.53 \\
    CDM~\cite{ho2022cascaded}  & 4.88 & - & 158.71 & - & - \\
    LDM-8~\cite{rombach2022high} & 15.51 & - & 79.03 & 0.65 & 0.63 \\
    LDM-8-G & 7.76 & - & 209.52 & 0.84 & 0.35 \\
    LDM-4 & 10.56 & - & 103.49 & 0.71 & 0.62 \\
    LDM-4-G & 3.60 & - & 247.67 &  {0.87} & 0.48 \\
    VDM++ \cite{kingma2023understanding} &2.12 &-&267.70 & - & -\\
    \multicolumn{6}{l}{ \textcolor{gray}{\emph{Diff. based on Transformer}}} \\
     U-ViT-H/2 \cite{bao2023all} &2.29 &5.68&263.88 &0.82 & 0.57 \\
     DiT-XL/2 \cite{peebles2023scalable}   &  {2.27} &  {4.60} &  {278.24} & 0.83 & 0.57 \\
     \multicolumn{6}{l}{ \textcolor{gray}{\emph{Diff. based on SSM}}} \\
     DiS-H/2  &2.10 &4.55 & 271.32 &0.82 & 0.58 \\
    \bottomrule
    \end{tabular}}
\end{table}

\subsubsection{Unconditional image generation.}

We conduct a comparative analysis between the DiT-S/2 model and prior diffusion models based on U-Net as well as Transformer-based diffusion models. 
Consistent with previous literature \cite{bao2023all}, we evaluate image quality using the FID score, based on 50K generated samples.
As summarized in Tables 2 and 3, DiS exhibits comparable performance to U-Net and U-ViT-S/2 on unconditional CIFAR10 and CelebA 64$\times$64 datasets, meanwhile significantly outperforming GenViT. Moreover, notably, our model achieves approximately a 50\% reduction in model parameters.

\subsubsection{Class-conditional image generation.}

To demonstrate performance in latent space, where images are initially converted to their latent representations before applying diffusion models, we conducted additional experiments. The evaluation results are listed in Tables 4 and 5. 
On the class-conditional ImageNet 256$\times$256 dataset, our DiS-H/2 achieves an FID of 2.10, surpassing all prior models. Notably, DiS-H/2 also outperforms DiT-XL/2, a competitive diffusion model employing a transformer as its backbone. Furthermore, on the class-conditional ImageNet 512$\times$512 dataset, our DiS-H/2 surpasses ADM-G, a model directly modeling image pixels.
Finally, we also observe that DiS-H/2 achieves higher recall values across all tested classifier-free guidance scales compared to LDM-4 and LDM-8 in various settings.

\subsection{Case Study}
\begin{figure}[t]
  \centering
   \includegraphics[width=0.96\linewidth]{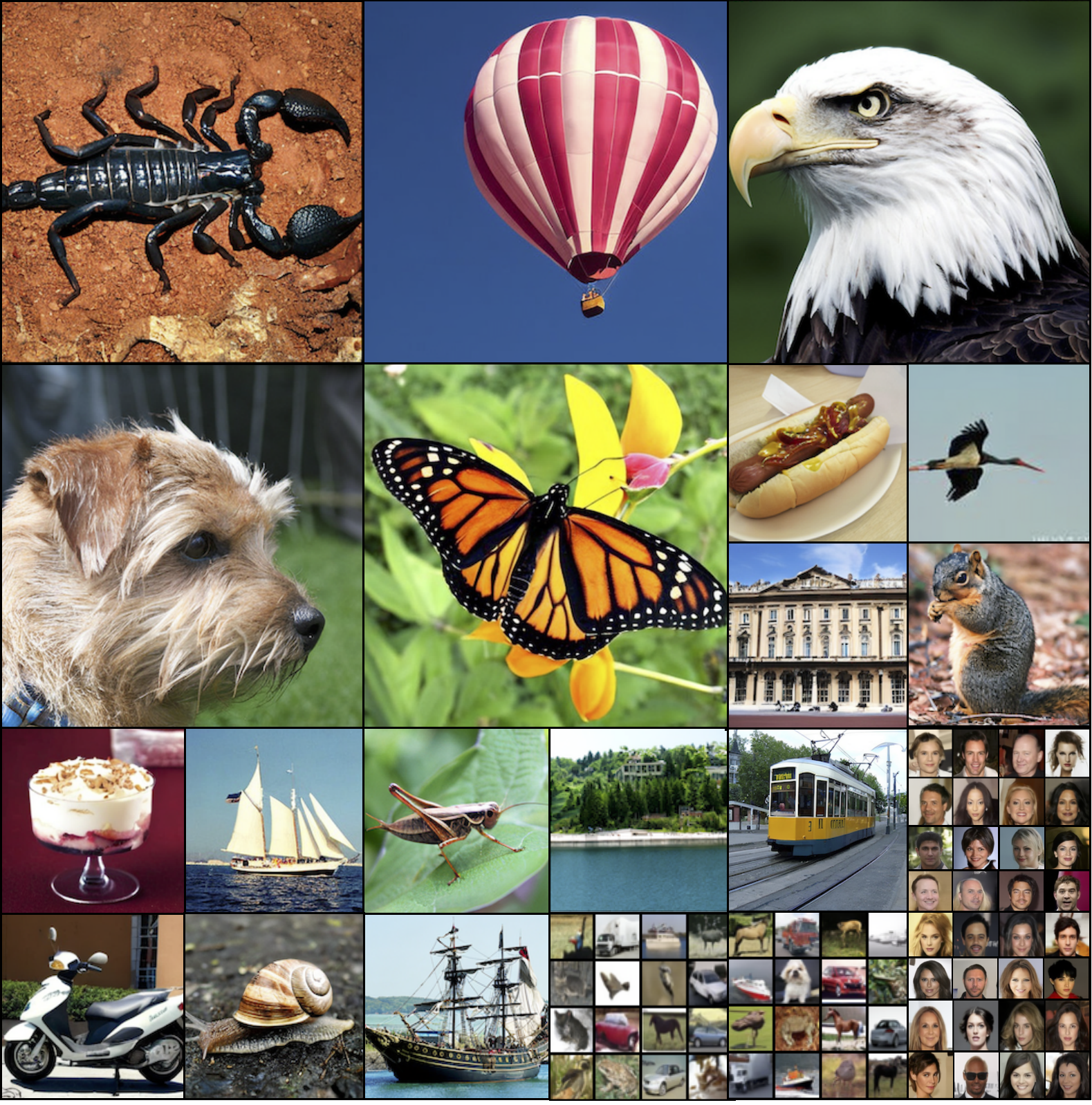}
   \caption{\textbf{Image results generated from DiS model. } Selected samples on ImageNet 256$\times$256, and random samples on CIFAR10, CelebA 64$\times$64. We can see that DiS can generate high-quality images while keeping integrated condition alignment. 
   }
   \label{fig:cases} 
\end{figure}

Figure \ref{fig:cases} showcases a selection of samples from ImageNet datasets at resolutions of 256$\times$256 and 512$\times$512, along with random samples from other datasets, demonstrating clear semantics and high-quality generation. For further exploration, additional generated samples, including both class-conditional and random ones, are available on the project page.

\begin{table}[t]
\centering
\caption{\textbf{Benchmarking class-conditional image generation on ImageNet 512$\times$512.} DiS-H/2 demonstrates a promising performance compared with both CNN-based and Transformer-based UNet for diffusion.} %
    \label{tbl:sota}
    \setlength{\tabcolsep}{2.5mm}{
    \begin{tabular}{lccccc}
    \toprule
    \multicolumn{6}{l}{\bf{Class-Conditional ImageNet} 512$\times$512} \\
    \midrule
    Model & FID$\downarrow$   & sFID$\downarrow$  & IS$\uparrow$     & Precision$\uparrow$ & Recall$\uparrow$ \\
     \midrule
     \multicolumn{6}{l}{ \textcolor{gray}{\emph{GAN}}} \\
    BigGAN-deep~\cite{brock2018large} & 8.43 & 8.13 & 177.90 & 0.88 & 0.29 \\
    StyleGAN-XL~\cite{sauer2022stylegan} & 2.41 & 4.06 & 267.75 & 0.77 & 0.52 \\
    \multicolumn{6}{l}{ \textcolor{gray}{\emph{Diff. based on U-Net}}}\\
    ADM~\cite{dhariwal2021diffusion} & 23.24 & 10.19 & 58.06 & 0.73 & 0.60  \\
    ADM-U & 9.96 & 5.62 & 121.78 & 0.75 &   {0.64} \\
    ADM-G & 7.72 & 6.57 & 172.71 &   {0.87} & 0.42 \\
    ADM-G, ADM-U & 3.85 & 5.86 & 221.72 & 0.84 & 0.53\\
    VDM++ \cite{kingma2023understanding} &2.65 &-&278.10 & - & -\\
    \multicolumn{6}{l}{ \textcolor{gray}{\emph{Diff. based on Transformer}}}\\
     U-ViT-H/4 \cite{bao2023all} &4.05 &6.44 &263.79&0.84 &0.48 \\
      {DiT-XL/2} \cite{peebles2023scalable} &   {3.04} &   {5.02} &   {240.82} & 0.84 & 0.54 \\
      \multicolumn{6}{l}{ \textcolor{gray}{\emph{Diff. based on SSM}}}\\
     DiS-H/2 & 2.88 &4.74  &272.33 & 0.84 &0.56 \\
    \bottomrule
    \end{tabular}}
\end{table}

\section{Related Works}

\subsection{State Space Backbone for Sequence Modeling}
State space models, a recent addition to the realm of deep learning, have garnered attention for their capacity to transform state spaces 
\cite{gu2021combining,gu2021efficiently,kim2021soft}. Drawing inspiration from continuous state space models in control systems and leveraging the HiPPO initialization \cite{gu2020hippo}, LSSL \cite{gu2021combining} has demonstrated promise in addressing long-range dependency issues. However, the computational and memory demands associated with the state representation render LSSL impractical for real-world applications. To mitigate this challenge, S4 \cite{gu2021efficiently} proposes normalizing parameters into a diagonal structure. Subsequently, various iterations of structured state-space models have emerged, featuring diverse structures such as complex-diagonal configurations  \cite{gupta2022diagonal,gu2022parameterization}, multiple-input multiple-output supporting \cite{smith2022simplified}, decomposition of diagonal plus low-rank operations \cite{hasani2022liquid}, selection mechanism \cite{gu2023mamba}. 
These models have been integrated into large representation models \cite{mehta2022long,ma2022mega,fu2022hungry}, primarily focusing on applications involving long-range and causal data such as language and speech, including tasks like language understanding \cite{ma2022mega} and content-based reasoning \cite{gu2023mamba}. 
Seminal work \cite{yan2023diffusion} is most related to us, they analyze attention-free structure for image synthesis. 
In this paper, we focus on the scaling properties of advanced Mamba structure when used as the backbone of diffusion models of images.

\subsection{Image Generation in Diffusions}

Diffusion models \cite{sohl2015deep,croitoru2023diffusion,dhariwal2021diffusion,ho2020denoising,nichol2021improved} represent a category of generative probabilistic models designed to approximate data distributions and facilitate straightforward sampling procedures. Typically, these models operate by taking a Gaussian noise input and iteratively denoising it through a series of gradual steps until it conforms to the target distribution. The specifics of the denoising process, including the number of steps and the transformation parameterization, vary across different studies \cite{sohl2015deep,song2020denoising,lu2022dpm,lu2022maximum,liu2022pseudo}. Recently, diffusion models have emerged as leading generators renowned for their ability to learn intricate distributions and produce diverse, high-quality samples. These models have found successful application across various domains, including images \cite{dhariwal2021diffusion,nichol2021improved,fei2022progressive,saharia2022photorealistic,ramesh2022hierarchical,rombach2022high,fei2023gradient,duan2023tuning}, video \cite{singer2022make}, 3D scenes \cite{muller2023diffrf}, motion sequences \cite{yuan2023physdiff,tevet2022human}, and so on. We follow the path of image generation with diffusion, but consider the impact of high-efficient structures.

\section{Conclusion}

This paper presents Diffusion State Space models (DiS),  a simple and general state space-based framework for image generation using diffusion models. 
DiS adopts a unified approach to handle all inputs, including time, conditions, and noisy image patches, treating them as concatenated tokens. Experimental results indicate that DiS compares favorably, if not surpasses, prior CNN-based or Transformer-based U-Net models while inheriting the remarkable scalability characteristic of the state space model class. We posit that DiS can offer valuable insights for future investigations into backbones within diffusion models and contribute to advancements in generative modeling across large-scale multimodal datasets. Given the encouraging scalability outcomes presented in this study, future endeavors should focus on further scaling DiS to larger models and token counts.

%
%
\bibliographystyle{splncs04}
\bibliography{main}

\begin{thebibliography}{10}
\providecommand{\url}[1]{\texttt{#1}}
\providecommand{\urlprefix}{URL }
\providecommand{\doi}[1]{https://doi.org/#1}

\bibitem{bao2023all}
Bao, F., Nie, S., Xue, K., Cao, Y., Li, C., Su, H., Zhu, J.: All are worth words: A vit backbone for diffusion models. In: Proceedings of the IEEE/CVF Conference on Computer Vision and Pattern Recognition. pp. 22669--22679 (2023)

\bibitem{betker2023improving}
Betker, J., Goh, G., Jing, L., Brooks, T., Wang, J., Li, L., Ouyang, L., Zhuang, J., Lee, J., Guo, Y., et~al.: Improving image generation with better captions. Computer Science. https://cdn. openai. com/papers/dall-e-3. pdf  \textbf{2}(3), ~8 (2023)

\bibitem{brock2018large}
Brock, A., Donahue, J., Simonyan, K.: Large scale gan training for high fidelity natural image synthesis. arXiv preprint arXiv:1809.11096  (2018)

\bibitem{cao2024survey}
Cao, H., Tan, C., Gao, Z., Xu, Y., Chen, G., Heng, P.A., Li, S.Z.: A survey on generative diffusion models. IEEE Transactions on Knowledge and Data Engineering  (2024)

\bibitem{chen2020wavegrad}
Chen, N., Zhang, Y., Zen, H., Weiss, R.J., Norouzi, M., Chan, W.: Wavegrad: Estimating gradients for waveform generation. arXiv preprint arXiv:2009.00713  (2020)

\bibitem{choi2021ilvr}
Choi, J., Kim, S., Jeong, Y., Gwon, Y., Yoon, S.: Ilvr: Conditioning method for denoising diffusion probabilistic models. arXiv preprint arXiv:2108.02938  (2021)

\bibitem{croitoru2023diffusion}
Croitoru, F.A., Hondru, V., Ionescu, R.T., Shah, M.: Diffusion models in vision: A survey. IEEE Transactions on Pattern Analysis and Machine Intelligence  (2023)

\bibitem{deng2009imagenet}
Deng, J., Dong, W., Socher, R., Li, L.J., Li, K., Fei-Fei, L.: Imagenet: A large-scale hierarchical image database. In: 2009 IEEE conference on computer vision and pattern recognition. pp. 248--255. Ieee (2009)

\bibitem{dhariwal2021diffusion}
Dhariwal, P., Nichol, A.: Diffusion models beat gans on image synthesis. Advances in neural information processing systems  \textbf{34},  8780--8794 (2021)

\bibitem{dockhorn2021score}
Dockhorn, T., Vahdat, A., Kreis, K.: Score-based generative modeling with critically-damped langevin diffusion. arXiv preprint arXiv:2112.07068  (2021)

\bibitem{dosovitskiy2020image}
Dosovitskiy, A., Beyer, L., Kolesnikov, A., Weissenborn, D., Zhai, X., Unterthiner, T., Dehghani, M., Minderer, M., Heigold, G., Gelly, S., et~al.: An image is worth 16x16 words: Transformers for image recognition at scale. arXiv preprint arXiv:2010.11929  (2020)

\bibitem{duan2023tuning}
Duan, X., Cui, S., Kang, G., Zhang, B., Fei, Z., Fan, M., Huang, J.: Tuning-free inversion-enhanced control for consistent image editing. arXiv preprint arXiv:2312.14611  (2023)

\bibitem{fei2023gradient}
Fei, Z., Fan, M., Huang, J.: Gradient-free textual inversion. arXiv preprint arXiv:2304.05818  (2023)

\bibitem{fei2022progressive}
Fei, Z., Fan, M., Huang, J., Wei, X., Wei, X.: Progressive denoising model for fine-grained text-to-image generation. arXiv preprint arXiv:2210.02291  (2022)

\bibitem{fu2022hungry}
Fu, D.Y., Dao, T., Saab, K.K., Thomas, A.W., Rudra, A., R{\'e}, C.: Hungry hungry hippos: Towards language modeling with state space models. arXiv preprint arXiv:2212.14052  (2022)

\bibitem{gu2023mamba}
Gu, A., Dao, T.: Mamba: Linear-time sequence modeling with selective state spaces. arXiv preprint arXiv:2312.00752  (2023)

\bibitem{gu2020hippo}
Gu, A., Dao, T., Ermon, S., Rudra, A., R{\'e}, C.: Hippo: Recurrent memory with optimal polynomial projections. Advances in neural information processing systems  \textbf{33},  1474--1487 (2020)

\bibitem{gu2022parameterization}
Gu, A., Goel, K., Gupta, A., R{\'e}, C.: On the parameterization and initialization of diagonal state space models. Advances in Neural Information Processing Systems  \textbf{35},  35971--35983 (2022)

\bibitem{gu2021efficiently}
Gu, A., Goel, K., R{\'e}, C.: Efficiently modeling long sequences with structured state spaces. arXiv preprint arXiv:2111.00396  (2021)

\bibitem{gu2021combining}
Gu, A., Johnson, I., Goel, K., Saab, K., Dao, T., Rudra, A., R{\'e}, C.: Combining recurrent, convolutional, and continuous-time models with linear state space layers. Advances in neural information processing systems  \textbf{34},  572--585 (2021)

\bibitem{gu2022vector}
Gu, S., Chen, D., Bao, J., Wen, F., Zhang, B., Chen, D., Yuan, L., Guo, B.: Vector quantized diffusion model for text-to-image synthesis. In: Proceedings of the IEEE/CVF Conference on Computer Vision and Pattern Recognition. pp. 10696--10706 (2022)

\bibitem{gupta2022diagonal}
Gupta, A., Gu, A., Berant, J.: Diagonal state spaces are as effective as structured state spaces. Advances in Neural Information Processing Systems  \textbf{35},  22982--22994 (2022)

\bibitem{hasani2022liquid}
Hasani, R., Lechner, M., Wang, T.H., Chahine, M., Amini, A., Rus, D.: Liquid structural state-space models. arXiv preprint arXiv:2209.12951  (2022)

\bibitem{heusel2017gans}
Heusel, M., Ramsauer, H., Unterthiner, T., Nessler, B., Hochreiter, S.: Gans trained by a two time-scale update rule converge to a local nash equilibrium. Advances in neural information processing systems  \textbf{30} (2017)

\bibitem{ho2022imagen}
Ho, J., Chan, W., Saharia, C., Whang, J., Gao, R., Gritsenko, A., Kingma, D.P., Poole, B., Norouzi, M., Fleet, D.J., et~al.: Imagen video: High definition video generation with diffusion models. arXiv preprint arXiv:2210.02303  (2022)

\bibitem{ho2020denoising}
Ho, J., Jain, A., Abbeel, P.: Denoising diffusion probabilistic models. Advances in neural information processing systems  \textbf{33},  6840--6851 (2020)

\bibitem{ho2022cascaded}
Ho, J., Saharia, C., Chan, W., Fleet, D.J., Norouzi, M., Salimans, T.: Cascaded diffusion models for high fidelity image generation. The Journal of Machine Learning Research  \textbf{23}(1),  2249--2281 (2022)

\bibitem{ho2022classifier}
Ho, J., Salimans, T.: Classifier-free diffusion guidance. arXiv preprint arXiv:2207.12598  (2022)

\bibitem{kalman1960new}
Kalman, R.E.: A new approach to linear filtering and prediction problems  (1960)

\bibitem{karras2022elucidating}
Karras, T., Aittala, M., Aila, T., Laine, S.: Elucidating the design space of diffusion-based generative models. Advances in Neural Information Processing Systems  \textbf{35},  26565--26577 (2022)

\bibitem{kim2021soft}
Kim, D., Shin, S., Song, K., Kang, W., Moon, I.C.: Soft truncation: A universal training technique of score-based diffusion model for high precision score estimation. arXiv preprint arXiv:2106.05527  (2021)

\bibitem{kingma2021variational}
Kingma, D., Salimans, T., Poole, B., Ho, J.: Variational diffusion models. Advances in neural information processing systems  \textbf{34},  21696--21707 (2021)

\bibitem{kingma2014adam}
Kingma, D.P., Ba, J.: Adam: A method for stochastic optimization. arXiv preprint arXiv:1412.6980  (2014)

\bibitem{kingma2023understanding}
Kingma, D.P., Gao, R.: Understanding diffusion objectives as the elbo with simple data augmentation. In: Thirty-seventh Conference on Neural Information Processing Systems (2023)

\bibitem{kingma2013auto}
Kingma, D.P., Welling, M.: Auto-encoding variational bayes. arXiv preprint arXiv:1312.6114  (2013)

\bibitem{kong2020diffwave}
Kong, Z., Ping, W., Huang, J., Zhao, K., Catanzaro, B.: Diffwave: A versatile diffusion model for audio synthesis. arXiv preprint arXiv:2009.09761  (2020)

\bibitem{krizhevsky2009learning}
Krizhevsky, A., Hinton, G., et~al.: Learning multiple layers of features from tiny images  (2009)

\bibitem{kynkaanniemi2019improved}
Kynk{\"a}{\"a}nniemi, T., Karras, T., Laine, S., Lehtinen, J., Aila, T.: Improved precision and recall metric for assessing generative models. Advances in Neural Information Processing Systems  \textbf{32} (2019)

\bibitem{liu2022pseudo}
Liu, L., Ren, Y., Lin, Z., Zhao, Z.: Pseudo numerical methods for diffusion models on manifolds. arXiv preprint arXiv:2202.09778  (2022)

\bibitem{liu2015deep}
Liu, Z., Luo, P., Wang, X., Tang, X.: Deep learning face attributes in the wild. In: Proceedings of the IEEE international conference on computer vision. pp. 3730--3738 (2015)

\bibitem{lu2022maximum}
Lu, C., Zheng, K., Bao, F., Chen, J., Li, C., Zhu, J.: Maximum likelihood training for score-based diffusion odes by high order denoising score matching. In: International Conference on Machine Learning. pp. 14429--14460. PMLR (2022)

\bibitem{lu2022dpm}
Lu, C., Zhou, Y., Bao, F., Chen, J., Li, C., Zhu, J.: Dpm-solver: A fast ode solver for diffusion probabilistic model sampling in around 10 steps. Advances in Neural Information Processing Systems  \textbf{35},  5775--5787 (2022)

\bibitem{ma2022mega}
Ma, X., Zhou, C., Kong, X., He, J., Gui, L., Neubig, G., May, J., Zettlemoyer, L.: Mega: moving average equipped gated attention. arXiv preprint arXiv:2209.10655  (2022)

\bibitem{mehta2022long}
Mehta, H., Gupta, A., Cutkosky, A., Neyshabur, B.: Long range language modeling via gated state spaces. arXiv preprint arXiv:2206.13947  (2022)

\bibitem{mei2023vidm}
Mei, K., Patel, V.: Vidm: Video implicit diffusion models. In: Proceedings of the AAAI Conference on Artificial Intelligence. vol.~37, pp. 9117--9125 (2023)

\bibitem{muller2023diffrf}
M{\"u}ller, N., Siddiqui, Y., Porzi, L., Bulo, S.R., Kontschieder, P., Nie{\ss}ner, M.: Diffrf: Rendering-guided 3d radiance field diffusion. In: Proceedings of the IEEE/CVF Conference on Computer Vision and Pattern Recognition. pp. 4328--4338 (2023)

\bibitem{nash2021generating}
Nash, C., Menick, J., Dieleman, S., Battaglia, P.W.: Generating images with sparse representations. arXiv preprint arXiv:2103.03841  (2021)

\bibitem{nichol2021improved}
Nichol, A.Q., Dhariwal, P.: Improved denoising diffusion probabilistic models. In: International Conference on Machine Learning. pp. 8162--8171. PMLR (2021)

\bibitem{orvieto2023resurrecting}
Orvieto, A., Smith, S.L., Gu, A., Fernando, A., Gulcehre, C., Pascanu, R., De, S.: Resurrecting recurrent neural networks for long sequences. arXiv preprint arXiv:2303.06349  (2023)

\bibitem{parmar2022aliased}
Parmar, G., Zhang, R., Zhu, J.Y.: On aliased resizing and surprising subtleties in gan evaluation. In: Proceedings of the IEEE/CVF Conference on Computer Vision and Pattern Recognition. pp. 11410--11420 (2022)

\bibitem{peebles2023scalable}
Peebles, W., Xie, S.: Scalable diffusion models with transformers. In: Proceedings of the IEEE/CVF International Conference on Computer Vision. pp. 4195--4205 (2023)

\bibitem{poole2022dreamfusion}
Poole, B., Jain, A., Barron, J.T., Mildenhall, B.: Dreamfusion: Text-to-3d using 2d diffusion. arXiv preprint arXiv:2209.14988  (2022)

\bibitem{ramesh2022hierarchical}
Ramesh, A., Dhariwal, P., Nichol, A., Chu, C., Chen, M.: Hierarchical text-conditional image generation with clip latents. arXiv preprint arXiv:2204.06125  \textbf{1}(2), ~3 (2022)

\bibitem{rombach2022high}
Rombach, R., Blattmann, A., Lorenz, D., Esser, P., Ommer, B.: High-resolution image synthesis with latent diffusion models. In: Proceedings of the IEEE/CVF conference on computer vision and pattern recognition. pp. 10684--10695 (2022)

\bibitem{saharia2022photorealistic}
Saharia, C., Chan, W., Saxena, S., Li, L., Whang, J., Denton, E.L., Ghasemipour, K., Gontijo~Lopes, R., Karagol~Ayan, B., Salimans, T., et~al.: Photorealistic text-to-image diffusion models with deep language understanding. Advances in Neural Information Processing Systems  \textbf{35},  36479--36494 (2022)

\bibitem{salimans2016improved}
Salimans, T., Goodfellow, I., Zaremba, W., Cheung, V., Radford, A., Chen, X.: Improved techniques for training gans. Advances in neural information processing systems  \textbf{29} (2016)

\bibitem{sauer2022stylegan}
Sauer, A., Schwarz, K., Geiger, A.: Stylegan-xl: Scaling stylegan to large diverse datasets. In: ACM SIGGRAPH 2022 conference proceedings. pp. 1--10 (2022)

\bibitem{singer2022make}
Singer, U., Polyak, A., Hayes, T., Yin, X., An, J., Zhang, S., Hu, Q., Yang, H., Ashual, O., Gafni, O., et~al.: Make-a-video: Text-to-video generation without text-video data. arXiv preprint arXiv:2209.14792  (2022)

\bibitem{smith2022simplified}
Smith, J.T., Warrington, A., Linderman, S.W.: Simplified state space layers for sequence modeling. arXiv preprint arXiv:2208.04933  (2022)

\bibitem{sohl2015deep}
Sohl-Dickstein, J., Weiss, E., Maheswaranathan, N., Ganguli, S.: Deep unsupervised learning using nonequilibrium thermodynamics. In: International conference on machine learning. pp. 2256--2265. PMLR (2015)

\bibitem{song2020denoising}
Song, J., Meng, C., Ermon, S.: Denoising diffusion implicit models. arXiv preprint arXiv:2010.02502  (2020)

\bibitem{song2021maximum}
Song, Y., Durkan, C., Murray, I., Ermon, S.: Maximum likelihood training of score-based diffusion models. Advances in Neural Information Processing Systems  \textbf{34},  1415--1428 (2021)

\bibitem{song2019generative}
Song, Y., Ermon, S.: Generative modeling by estimating gradients of the data distribution. Advances in neural information processing systems  \textbf{32} (2019)

\bibitem{song2020score}
Song, Y., Sohl-Dickstein, J., Kingma, D.P., Kumar, A., Ermon, S., Poole, B.: Score-based generative modeling through stochastic differential equations. arXiv preprint arXiv:2011.13456  (2020)

\bibitem{tevet2022human}
Tevet, G., Raab, S., Gordon, B., Shafir, Y., Cohen-Or, D., Bermano, A.H.: Human motion diffusion model. arXiv preprint arXiv:2209.14916  (2022)

\bibitem{vahdat2021score}
Vahdat, A., Kreis, K., Kautz, J.: Score-based generative modeling in latent space. Advances in Neural Information Processing Systems  \textbf{34},  11287--11302 (2021)

\bibitem{yan2023diffusion}
Yan, J.N., Gu, J., Rush, A.M.: Diffusion models without attention. arXiv preprint arXiv:2311.18257  (2023)

\bibitem{yang2022your}
Yang, X., Shih, S.M., Fu, Y., Zhao, X., Ji, S.: Your vit is secretly a hybrid discriminative-generative diffusion model. arXiv preprint arXiv:2208.07791  (2022)

\bibitem{yuan2023physdiff}
Yuan, Y., Song, J., Iqbal, U., Vahdat, A., Kautz, J.: Physdiff: Physics-guided human motion diffusion model. In: Proceedings of the IEEE/CVF International Conference on Computer Vision. pp. 16010--16021 (2023)

\bibitem{zhang2023adding}
Zhang, L., Rao, A., Agrawala, M.: Adding conditional control to text-to-image diffusion models. In: Proceedings of the IEEE/CVF International Conference on Computer Vision. pp. 3836--3847 (2023)

\bibitem{zhao2022egsde}
Zhao, M., Bao, F., Li, C., Zhu, J.: Egsde: Unpaired image-to-image translation via energy-guided stochastic differential equations. Advances in Neural Information Processing Systems  \textbf{35},  3609--3623 (2022)

\bibitem{zhou2023sparsefusion}
Zhou, Z., Tulsiani, S.: Sparsefusion: Distilling view-conditioned diffusion for 3d reconstruction. In: Proceedings of the IEEE/CVF Conference on Computer Vision and Pattern Recognition. pp. 12588--12597 (2023)

\bibitem{zhu2024vision}
Zhu, L., Liao, B., Zhang, Q., Wang, X., Liu, W., Wang, X.: Vision mamba: Efficient visual representation learning with bidirectional state space model. arXiv preprint arXiv:2401.09417  (2024)

\end{thebibliography}
\end{document}